\documentclass[letterpaper, 10 pt, conference]{ieeeconf}
\IEEEoverridecommandlockouts 
\usepackage{graphicx}                       
\usepackage{graphics}                       
\usepackage{epsfig}                         
\usepackage{caption}
\usepackage{subcaption} 

\usepackage{amssymb,amsmath}    
\DeclareMathOperator*{\argmax}{arg\,max}

\usepackage{mdwmath}
\usepackage{commath}   
\usepackage{eqparbox}
\usepackage{mathtools}
\usepackage[utf8]{inputenc} 
\usepackage[english]{babel}
\usepackage{amsmath,amsfonts,amssymb}

\newcommand{\beq}{\begin{equation}}
\newcommand{\eeq}{\end{equation}}
\newcommand{\bear}{\begin{eqnarray}}
\newcommand{\bears}{\begin{eqnarray*}}
\newcommand{\eear}{\end{eqnarray}}
\newcommand{\eears}{\end{eqnarray*}}
\newcommand{\bdm}{\begin{displaymath}}
\newcommand{\edm}{\end{displaymath}}
\newcommand{\lba}{\left[\begin{array}}
\newcommand{\ear}{\end{array}\right]}

\usepackage{stfloats}                       
\usepackage{url} 
\usepackage{hyperref}
\usepackage{cite}                           
\usepackage[T1]{fontenc} 
\usepackage{color}
\usepackage{xcolor}
\definecolor{forestgreen}{rgb}{0.13, 0.55, 0.13}
\usepackage{algorithm}
\usepackage[noend]{algpseudocode}

\usepackage{authblk}
\title{\LARGE \bf
Hyperparameter Auto-tuning in Self-Supervised Robotic Learning
}

\author{Jiancong Huang$^{\ast1}$, Juan Rojas$^{\ast2}$, Matthieu Zimmer$^{3}$, Hongmin Wu$^{4}$, Yisheng Guan$^{1}$, and Paul Weng$^{3,5}$%
\thanks{$^{1}$School of Electromechanical Engineering, Guangdong University of Technology; $^{2} $Department of Mechanical and Automation Engineering, The Chinese University of Hong Kong; $^{3} $UM-SJTU Joint Institute, Shanghai Jiao Tong University; $^{4} $Institute of Intelligent Manufacturing, Guangdong Academy of Sciences; $^5 $Department of Automation, Shanghai Jiao Tong University; $\ast$Authors contributed equally.}
\thanks{Corresponding author: \url{paul.weng@sjtu.edu.cn}}
\thanks{
This paper was published in \href{https://ieeexplore.ieee.org/document/9372868}{IEEE Robotics and Automation Letters (RA-L)}, presented at ICRA 2021, and presented at NeurIPS 2020 DRL workshop.
}
\thanks{Digital Object Identifier (DOI): 10.1109/LRA.2021.3064509}
}

\begin{document}
\maketitle 
\thispagestyle{empty} 
\pagestyle{empty} 
\begin{abstract}
Policy optimization in reinforcement learning requires the selection of numerous hyperparameters across different environments. 
Fixing them incorrectly may negatively impact optimization performance leading notably to insufficient or redundant learning.
Insufficient learning (due to convergence to local optima) results in under-performing policies whilst redundant learning wastes time and resources. 
The effects are further exacerbated when using single policies to solve multi-task learning problems.
Observing that the Evidence Lower Bound (ELBO) used in Variational Auto-Encoders correlates with the diversity of image samples, we propose an auto-tuning technique based on the ELBO for self-supervised reinforcement learning. 
Our approach can auto-tune three hyperparameters: the replay buffer size, the number of policy gradient updates during each epoch, and the number of exploration steps during each epoch.
We use a state-of-the-art self-supervised robot learning framework (Reinforcement Learning with Imagined Goals (RIG) using Soft Actor-Critic) as baseline for experimental verification.
Experiments show that our method can auto-tune 
online and yields the best performance at a fraction of the time and computational resources.
Code, video, and appendix for simulated and real-robot experiments can be found at the project page \url{www.JuanRojas.net/autotune}.
\end{abstract}


\section{Introduction}\label{sec:introduction}
Most Reinforcement Learning (RL) algorithms depend on many hyperparameters and their performance is very sensitive to them \cite{Henderson2018DeepRL,closer_look_at_pi_g}.
Hyperparameter tuning is traditionally performed via an outer loop with methods such as grid search, random search \cite{bergstra2012random}, or Bayesian optimization \cite{akiba2019optuna}.
Tuning requires extensive computing resources and is challenging in practice, especially with curriculum learning or in multi-task settings since good hyperparameter values may be task-dependent.
In contrast, some works in RL try to tune hyperparameters online in a single run, such as learning rates \cite{Gupta2019FiniteTimePB}, the bootstrapping parameter $\lambda$ in $\lambda$-returns \cite{White2016AGA}, or any differentiable hyperparameters \cite{zahavy2020self}. 

Auto-tuning hyperparameters in a single run is appealing as it is much more efficient than traditional methods, however it is also much more challenging to achieve.
The poor optimization of any hyperparameters can negatively impact an agent's learning progress cumulatively. 
For instance, very small or large hyperparameter values may slow down convergence, or even lead to divergence.
Even if it converges, the trained policy may be suboptimal and a slower convergence results in a waste of computational, space, and sample resources, the latter being particularly costly in robotics.

Our work follows this promising line of research and proposes a novel auto-tuning method, which we demonstrate in the Reinforcement Learning  with Imagined Goals (RIG) algorithm \cite{nair2018visual, nair19ccrig,pong2019skew}. 
We expect that our auto-tuning technique could also be adapted to other deep RL algorithms based on Variational Auto-Encoders (VAE) and using visual inputs, but we leave this for future work.

In self-supervised reinforcement learning, RIG \cite{nair2018visual, nair19ccrig,pong2019skew} learns general-purpose skills in task-agnostic environments using only RGB images as observations without a reward function designed for each task.
RIG leverages self-generated goals to collect vast skills from the environment. 
In RIG, an agent learns to reach goals produced by a VAE-based goal generator that is fine-tuned (online) with collected samples stored in a replay buffer.

In this paper, we propose to use the value of the negative ELBO, which is the optimized loss function in VAEs, to estimate the number of different goals.
Based on this premise, our method auto-tunes three hyperparameters to optimize learning in VAE-based environments; namely:
\begin{itemize}
  \item the number $\mathbf{N_e}$ of exploration steps in each epoch for sampling goals and interacting with the environment with a sufficient number of steps in each epoch.
  \item the replay buffer size $\mathbf{N_b}$ so as not to lose (or forget) transitions nor waste memory resources.
  \item the number $\mathbf{N_{\theta}}$ of gradient updates in each epoch to optimally update the policy at each epoch.
\end{itemize}

Our proposition is motivated by the following observations: (i)
the value of the negative $\mathrm{ELBO}$ is positively correlated with the diversity in the training samples and thus also positively correlated with the number $\mathbf{N}_{\mathbf{g}}$ of different goals in the replay buffer for RIG, and
(ii) the hyperparameters $\mathbf{N}_{\mathbf{e, b,\theta}} = (\mathbf{N_e}, \mathbf{N_b}, \mathbf{N_{\theta}})$ can be approximated or upperbounded by a multiple of $\mathbf{N}_{\mathbf{g}}$ (see Sections~\ref{sec:choice} and \ref{sec:diversity_of_r_b} for a detailed discussion).
Therefore, the negative $\mathrm{ELBO}$ can be used to auto-tune these three hyperparameters.

The contributions of our work are threefold:
(i) we identify that the loss function of VAEs is related to the diversity of the samples; 
(ii) to avoid suboptimal learning or wasted computer or sample resources, we propose a methodology that auto-tunes three hyperparameters $\mathbf{N}_{\mathbf{e,b,\theta}}$;
(iii) we experimentally validate our approach on diverse domains, and we additionally report competitive performances in curriculum learning settings.
\section{Background}\label{sec:background}
\subsection{Visual Reinforcement Learning with Imagined Goals. }
In RL, given a state observation $s \in \mathcal{S}$, an agent decides which action $a \in \mathcal{A}$ to take, where $\mathcal{S}$ is the state observation space and $\mathcal{A}$ is the action space. 
To quantify the goodness of an agent's behavior, a reward function $R : \mathcal{S} \to \mathbb{R}$ yields a reward signal at each time step depending on the observation. 
Hence, the agent has to learn a policy $\pi : \mathcal{S} \to \Delta(\mathcal{A})$ to maximize an expected cumulative reward, where $\Delta(\mathcal{A})$ denotes the set of probability measures over $\mathcal{A}$. 

In goal-conditioned RL \cite{pmlr-v37-schaul15}, the policy must also take a goal $g \in \mathcal{G}$ into account, where $\mathcal{G}$ is the goal space. The policy definition becomes $\pi : \mathcal{S} \times \mathcal{G} \to \Delta(\mathcal{A})$ and the reward function $R : \mathcal{S} \times \mathcal{G} \to \mathbb{R}$ indicates the proximity to the goal. 

In self-generated goal RL, goals are generated by the agent instead of being provided by the environment.
More specifically, in RIG \cite{nair2018visual}, an observation and a goal belong to the same space $\mathcal{S} = \mathcal{G}$, they represent the current observed image and the target image respectively.
To simplify the problem, instead of working in the high-dimensional observation space, all the input is compressed into a latent space $\mathcal{Z}$ via an encoder $p_{\phi} : \mathcal{S} \to \Delta(\mathcal{Z})$ with parameters $\phi$. 
As the observation and goal spaces are equivalent, both can be encoded.
Given an observation $s \in \mathcal{S}$ and a goal $g \in \mathcal{G}$, we denote $z_s \sim p_{\phi}(z|s) $ and $z_g \sim p_{\phi}(z|g)$ as their associated latent representations.
Therefore, the policy can be decomposed as $\pi(s, g) = \pi_\theta (z_s , z_g) $ with $\pi_\theta : \mathcal{Z} \times \mathcal{Z} \to \Delta(\mathcal{A})$ being a parametric policy defined in the latent space where $\theta$ represents the parameters (weights of a neural network) to be learned.
They also assume that the reward function 
is known to the agent and defined over the latent space as follows: $r(z_s, z_g) = -||z_s - z_g||_2$. 
In this paper, as in \cite{nair2018visual}, we use the L2 norm instead of the Mahalanobis distance as it was shown to work better empirically.
Thus, the best parameters $\theta^*$ for the policy are defined as:
\begin{equation}\label{eq:g_condi_rl}
	\theta^* = \argmax_{\theta}\ \mathbb{E}_{z_g \sim \mathcal{N}(0,\mathrm{I})} \Big[ \mathbb{E}_{\pi_{\theta}}\big[ \sum_t \gamma^t r(z^t_s, z_g) \big] \Big],
\end{equation}
where $\mathrm{I}$ is an identity matrix with equal dimension as $\mathcal{Z}$,  self-generated goals $z_g$ are acquired by sampling a unit Gaussian, and $\gamma \in [0, 1)$ is a discount factor.

\subsection{Variational Auto-Encoders and Evidence Lower Bound. }\label{sec:ELBO_change_on_fixed_H}
In VAEs, the goal is to approximate the intractable $p(z|x)$ with a parametric function $p_{\phi}(z|x)$ by optimizing an objective function expressed with $\mathrm{KL}(p_{\phi}(z|x)||p(z|x))$. 
Optimizing this objective function is equivalent to maximizing the Evidence Lower Bound (ELBO):
\begin{equation}\label{eq:elbo_and_vae_4}
	\begin{split}
	& -\mathrm{KL}(p_{\phi}(z|x)||p(z|x)) \\
	& \geq  \underbrace{\mathbb{E}_{p_{\phi}(z|x)} \log p_{\psi}(x|z) - \mathrm{KL} (p_{\phi}(z|x) || q(z))}_{\mathrm{Evidence\ Lower\ BOund\ (ELBO)}}.
	\end{split}
\end{equation}
where $q(z) = \mathcal{N}(0, \mathbf{I})$.
The loss function of VAEs is the negative $\mathrm{ELBO}$\cite{diederik2014auto}\cite{rezende2014stochastic}\cite{alemi2018fixing}.
Minimizing the loss function amounts to minimizing $\mathrm{KL}(p_{\phi}(z|x)||p(z|x))$, as shown by Eq. \ref{eq:elbo_and_vae_4}. 
Finally, combining with $\beta \geq 1$, we obtain the $\beta$-Variational Auto-Encoder's ($\beta$-VAE) loss function \cite{higgins2017beta}, which we call $\beta$-ELBO, or simply ELBO when there is no risk of confusion:
\begin{equation}\label{eq:negative_evidence_lower_bound}
	\begin{split}
        \mathcal{L}_{\psi,\phi}
        = \beta \mathrm{KL}(p_{\phi}(z|x)||q(z)) - \mathbb{E}_{p_{\phi}(z|x)} \log p_{\psi}(x|z).
	\end{split}
\end{equation}

The term $\beta \mathrm{KL}(p_{\phi}(z|x)||q(z))$ is the Kullback–Leibler divergence between $p_{\phi}(z|x)$ and $q(z)$, with hyperparameter $\beta$ controlling the trade-off between reconstruction quality and disentanglement. 
The second term $\mathbb{E}_{p_{\phi}(z|x)} \log p_{\psi}(x|z)$ is the reconstruction log-probability.
The loss function in Eq. \ref{eq:negative_evidence_lower_bound} is minimized by simultaneously optimizing with respect to $\phi$ and $\psi$.
In summary, it is through this process that $\beta$-VAE learns a latent representation that is maximally expressive about reconstruction and maximally compressive about the samples.
\subsection{The Entropy of Observations and Goals. }\label{subsec:it_and_goal_generator}
Let $X = \{x_1, \ldots, x_n\}$ be a dataset of $n$ samples.
This dataset induces an empirical distribution where $p(x)$ is the probability of sampling $x$ from $X$.
Then the (empirical) entropy of $X$ is $\mathrm{H}(X) = - \sum_{k=1}^{n} p(x_k) \log p(x_{k}) \leq \log(n)$.
The inequality holds when all the samples are different.
In that case, the empirical distribution becomes the uniform distribution.
The entropy is maximum when all the samples are different. 
In other words, maximum entropy happens when there is maximal diversity in the dataset. 
More generally, if the samples are obtained from some true distribution $p(x)$, then as the number of samples $n$ increases, $\mathrm{H}(X)$ tends to the entropy of $p(x)$.
In our work, the samples are the visual observations generated in RIG. 
We use the (empirical) entropy to measure its diversity.

\subsection{The Problem of Choosing $\mathbf{N}_{\mathbf{e,b,\theta}}$.} \label{sec:choice}
In RIG \cite{nair2018visual}, the training progress is composed of sequential epochs. 
Each and every epoch has an exploration phase, a policy update phase, a policy evaluation phase, and a VAE fine-tuning phase as delineated in Algo. \ref{alg:algorithm}. 
RIG uses fixed values for the hyperparameters $\mathbf{N}_{\mathbf{e,b,\theta}}$. 
In other words, they need to be pre-tuned before running RIG.

The number of exploration steps, $\mathbf{N_e}$, specifies how many times a goal is generated.
For each goal, the agent interacts with the environment and stores transitions into the replay buffer. 
In curriculum learning, different environments may need different number of exploration steps at each epoch since the diversity of their goals is different. 
A sufficiently large value for $\mathbf{N_{e}}$ ensures that the agent interacts sufficiently with all possible goals in an environment to get sufficient novel transitions for policy updates.
If $\mathbf{N_e}$ is too small, the current policy $\pi_{\theta i}$ would be updated with only a limited number of collected transitions, which may lead to a sub-optimal policy.
On the other hand, if $\mathbf{N_e}$ is excessive, the redundant steps will be uninformative but costly in space, time and samples, which is particularly prohibitive in real-world setups.
As is known, high time and sample complexity is an outstanding problem in DRL and one that prevents its wider adoption in real-world scenarios. 
The value of $\mathbf{N_e}$ should be such that the agent interacts with all the possible goals at least once.

The replay buffer size, $\mathbf{N_b}$, controls how many transitions $\tau=(s, a, s', z_g)$ are saved in the replay buffer $\mathcal{B}$ from the exploration phase. 
If the size of the replay buffer $\mathbf{N_b}$ is insufficient, the buffer will be filled with transitions very quickly. 
Then, the stored transitions would be replaced by novel ones leading to forgotten behaviors from previous exploration steps.
On the other hand, if $\mathbf{N_b}$ is too large, unnecessary space is used to maintain the large buffer. The value of $\mathbf{N_b}$ should be at least equal to the product of $\mathbf{N_e}$ and the trajectory length $l$ to be able to store all the samples that are generated during those exploration steps.

The number of gradient updates, $\mathbf{N_\theta}$, defines how many times the policy $\pi_{\theta}$ is updated by sampling transitions from the replay buffer and using the future strategy \cite{nair2018visual}.  
If $\mathbf{N_{\theta}}$ is too small or too large, the learning of the agent may be hindered.
In the best case, the convergence would be slower and in the worst case, the process would diverge.
Besides, if $\mathbf{N_{\theta}}$ is too large, even if the agent does manage to learn, resources in terms of time, computation, space, and samples would be unnecessarily consumed.
The value of $\mathbf{N_\theta}$ should be such that we could sample all the different transitions for gradient update at least once.

\section{Auto-tuning the Hyperparameters} \label{sec:autotune_scaffolding_hyperparameters}
\subsection{Estimation of Samples Diversity with ELBO.} 
\label{subsec:elbo_and_diversity}
We show how the diversity of a dataset $X$, as measured by its (empirical) entropy $\mathrm{H(\mathit{X})}$, can be related to the ELBO.
We denote $\mathrm{I}(X;\mathcal{Z})$ as the (empirical) mutual information between samples in $X$ and in the latent space $\mathcal Z$. 
By abuse of notations, we identify the random variables corresponding to those sampling distributions with the corresponding spaces (i.e., $X$ and $\mathcal Z$).
Formally, we have:
\begin{align}
	\mathrm{I(\mathit{X};\mathcal{Z})} &= \mathrm{H(\mathit{X})} - \mathrm{H(\mathit{X}|\mathcal{Z})} \nonumber\\
    = & \mathbb{E}_{\hat p(x,z)}\left[\log \frac{\hat p(z|x)}{\hat p(z)}\right] 
\nonumber\\
    \approx & \mathbb{E}_{p(x,z)}\left[\log \frac{p(z|x)}{p(z)}\right] \leq \mathbb{E}_{p(x,z)}\left[\log \frac{p(z|x)}{q(z)}\right], \label{eq:upper_bound_of_MI1}
\end{align}
where the distributions $\hat p$'s are induced by the sampling distribution over $X$, $p(z)$ is the prior distribution over latent factor $z$, and $q(z) = \mathcal N(0, \mathrm{I})$.
Note that as the size of the dataset tends to infinity, the approximation error in \eqref{eq:upper_bound_of_MI1} would vanish.
By isolating $\mathrm{H}(X)$, we obtain:
\begin{align}
\mathrm{H(\mathit{X})} & \mathrel{\lessapprox} \mathbb{E}_{p(x,z)}\left[\log \frac{p(z|x)}{q(z)}\right] + \mathrm{H(\mathit{X}|\mathcal{Z})} \nonumber\\
&\mathrel{\lessapprox} \mathbb{E}_{p(x,z)}\left[\log \frac{p(z|x)}{q(z)}\right] - \mathbb{E}_{p(x,z)}\left[\log p(x|z)\right] \nonumber\\
&\mathrel{\lessapprox} \mathbb{E}_{p(x)} \left[ \mathrm{KL}(p(z|x)||q(z)) - \mathbb{E}_{p(z|x)} \log p(x|z) \right] \nonumber\\
& \lessapprox \mathbb{E}_{p(x)}\left[-\mathrm{ELBO}\right] \label{eq:ELBOupperbound}\\
 	&\lessapprox
 	\mathbb{E}_{p(x)} \left[ \beta \mathrm{KL}(p_{\phi}(z|x)||q(z)) - \mathbb{E}_{p_{\phi}(z|x)} \log p_{\psi}(x|z) \right] \nonumber\\
&\lessapprox
\mathbb{E}_{p(x)} \left[ -\beta\mbox{-}\mathrm{ELBO} \right]. \label{eq:betaELBOupperbound}
\end{align}

Eq.~\ref{eq:ELBOupperbound} holds because in the context of VAE, $p(z|x)$ and $p(x|z)$ are approximated by $p_{\phi}(z|x)$ and $p_{\psi}(x|z)$. 
Eq.~\ref{eq:betaELBOupperbound} holds because $\beta\ge 1$ in $\beta$-VAE and KL divergences are positive.


The learning process of $\beta$-VAEs as described in Sec. \ref{sec:ELBO_change_on_fixed_H} is performed via gradient descent, which attempts to drive the loss to the lower bound $\mathrm{H}(\mathit{X})$. 
From Eq. \ref{eq:betaELBOupperbound}, it is possible to explain why the $\beta$-ELBO tends to stable values during training when the diversity of the samples does not change.
At convergence, the right-hand side of Eq. \ref{eq:betaELBOupperbound} provides a good approximation of the diversity of the dataset.
Furthermore, Eq. \ref{eq:betaELBOupperbound} suggests a potential interesting relation between $\mathrm{H}(\mathit{X})$, the expected KL divergence and the expected log-probability, which we observe experimentally in some simple conditions (see Fig.~\ref{fig:explain_kl_change}):
when the diversity increases, the KL divergence term would tend to increase and the log-probability term to decrease, and vice versa.

\subsection{Auto-tuning of the Hyperparameters. }\label{sec:diversity_of_r_b} 
In a VAE-based goal generator, the next observations $s'$ of transitions $(s, a, s')$ stored in a replay buffer $\mathcal B$ are used to fine-tune the goal generator. 
Let $\mathcal{B}'$ denote the dataset of next observations induced by $\mathcal{B}$.
In Sec. \ref{subsec:it_and_goal_generator}, the number $\mathbf{N_{s'}}$ of next observations in $\mathcal{B}'$ is related to its entropy:  $\mathrm{H}(\mathcal{B}') \le \log(\mathbf{N_{s'}})$.
For simplicity, we use an approximate linear lowerbound: $\mathbf{N_{s'}} \gtrapprox \xi \mathrm{H}(\mathcal{B}')$, where scaling factor $\xi$ is a new hyperparameter that we introduce.
Therefore, the number of different observations can be approximately lower-bounded as follows: $\mathbf{N_{s'}} \gtrapprox \xi \mathrm{H}(\mathcal{B}') \gtrapprox -\xi \beta\mbox{-}\mathrm{ELBO}$. 
In self-supervised RL, the number of goals $\mathbf{N_g}$ is upper-bounded by the number of  observations $\mathbf{N_{s'}}$.
Therefore, we can approximate the number of goals with $-\xi\beta\mbox{-}\mathrm{ELBO}$ by choosing $\xi$ correctly because $\mathbf{N_g}$ and $-\xi\beta\mbox{-} \mathrm{ELBO}$ both lower-bound $\mathbf{N_{s'}}$.
As a side note, during the policy learning phase in RIG, the latent representations of the observations and the goals are actually used instead of the high dimensional images collected from the environment.

Given the rationale of the relations between the three hyperparameters $\mathbf{N}_{\mathbf{e,b,\theta}}$ in Sec. \ref{sec:choice}, we can further establish a few more relations to develop heuristics for auto-tuning $\mathbf{N}_{\mathbf{e,b,\theta}}$. 
For the number of exploration steps, recall that an agent takes exploration steps to generate goals and interact with the environment. To guarantee that the agent collects sufficient transitions with different goals, the least number of exploration steps should be defined by $\mathbf{N}_{\mathbf{e}} \geq \mathbf{N_g}$. 
%
For the replay buffer size, the minimum size needed is $\mathbf{N_b} \geq l \cdot \mathbf{N}_{\mathbf{e}} $, where $l$ is the maximum path length of a trajectory during each exploration step.
%
For the number of policy gradient updates, 
instead of calculating the number of different transitions, the least number of steps an agent would take is lower bounded by the number of different goals: $\mathbf{N}_{\mathbf{\theta}} \geq \mathbf{N_g}$. 
 

In conclusion, given 
the previous discussion about
the relations between next states, goals, exploration steps, buffer size, and gradient updates, we propose an auto-tuning algorithm in RIG (Algo. \ref{alg:algorithm}) where a single hyperparameter $\xi$ for scaling the negative $\beta\mbox{-}\mathrm{ELBO}$ replaces $\mathbf{N_{e,b,\theta}}$ (see line~2).

\begin{algorithm}[tbh]
    \begin{algorithmic}[1]
        \caption{Auto-tuning $\mathbf{N}_{\mathbf{e,b,\theta}}$ in RIG}
        \label{alg:algorithm}
        \Require
        \Statex 
        Policy $\pi_{\theta_{0}}$ from an off-policy RL algorithm.       
        
        \Statex
        Max path length $l$ until done, and scaling factor $\xi$.        
        
        \Statex 
        Initialize $\mathbf{N}_{\mathbf{b}0, \mathrm{\theta}0, \mathbf{e}0}$, encoder $p_{\phi_{0}}$, decoder $p_{\psi_{0}}$, $\mathcal{L}_{\psi_{0}, \phi_{0}}$, replay buffer $\mathcal{B}_{0}$, $\mathcal{B}^{s'}_{0}$ refers to all the next observations $\mathcal{S'}$ in $\mathcal{B}_{0}$.

        
        \For{Epoch $i=1, \ldots, E$}
        \State $\mathbf{N}_{\mathbf{b}i} = -\xi \mathrm{ELBO} l $, $\mathbf{N}_{\mathbf{\theta}i} = \mathbf{N}_{\mathbf{e}i} = -\xi \mathrm{ELBO}$.   \Comment{auto-tuning, $\mathbf{N}_{{\mathbf{b}i,\mathbf{e}i,\mathbf{\theta}i}} \geq 1$}
        
        \For{Exploration steps $n=0, \ldots, \mathbf{N}_{\mathbf{e}i} $}
        \State Sample latent goal.
        \State Interact with environment following $\pi_{\theta i}$ and get path $(s_0, a_0, s_1, ..., s_{t-1}, a_{t-1}, s_{t}, z_g)$, where $t \leq l$.
        \State Encode all image observation $s$ with $p_{\phi_{i}}$.
        \State Store episode into $\mathcal{B}_{i}$, possibly remove older episodes to ensure size $|\mathcal{B}_{i}| \le \mathbf{N}_{\mathbf{b}i}$.
        \EndFor

        \For{Policy updates number $n=0, \ldots,$ $\mathbf{N}_{\mathbf{\theta}i}$}
        \State Sample batch from $\mathcal{B}_{i}$ to update $\pi_{\theta i}$.
        \EndFor
        \State Policy evaluation with specific goals.
        \If{Fine-tune $\beta$-VAE}
        \State Sample batch from $\mathcal{B}_{i}$ to update $p_{\phi i}$, $p_{\psi i}$.
        \State Evaluate $p_{\phi i}$, $p_{\psi i}$. 
        \State Compute $\mathrm{ELBO}$.
        \EndIf
        \EndFor
    \end{algorithmic}
\end{algorithm}
\section{Experiments}\label{sec:experiments}
In this section, 
we first demonstrate the relation between the ELBO and the diversity of the corresponding dataset--if the data is representative, we can also consider it as the diversity of the environment as described in Sec. \ref{subsec:elbo_and_diversity}.
Then we test and analyze the performance of the hyperparameter optimization of $\mathbf{N_{e,b,\theta}}$ and $\xi$.
We also visualize the quality of the goals coverage using our auto-tuning method and compare it to the baselines.
Finally, we compare the performance of our auto-tuning method in curriculum learning environments.
\begin{figure*}[!h]
  \centering
    \includegraphics[width=0.95\linewidth]{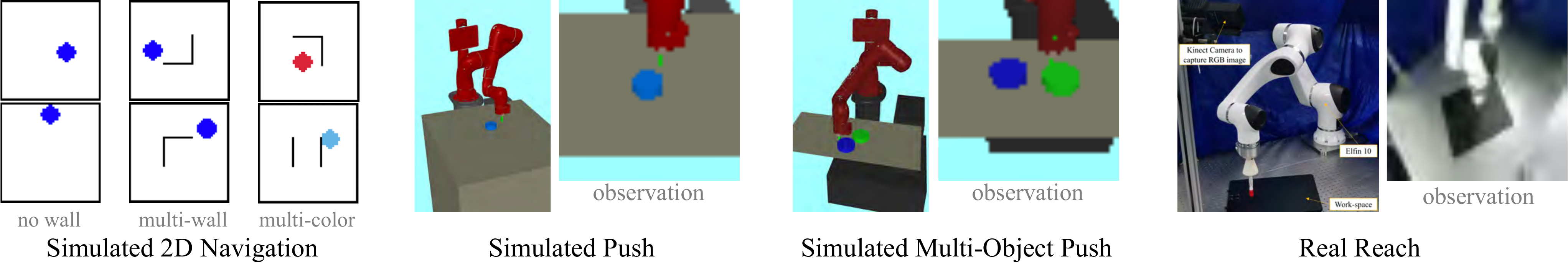}
    \caption{The simulated and real-robot environments. In both cases, the $3 \times 48 \times 48$ RGB image observations are flattened and normalized into $6912 \times 1$ as the state $s$ for the RIG framework. Actions are defined as the 2D positions of the point in the navigation environment and the 3D positions of the end-effector in the robot environment. In 2D navigation, the goal is to navigate a point. The wall is randomly chosen from a set of 15 possible configurations in each exploration step, and and the color of the point is a randomly sampled RGB value. In the robot environment, an arm is tasked with pushing an object to a target position or reaching a target position.}
  \label{fig:screenshot_and_obs}
\end{figure*}
\subsection{Does sample diversity change the ELBO? }\label{exp:1_verify_on_mnist_and_visual_robot}
\begin{figure}[]
  \centering
    \includegraphics[width=.95\linewidth]{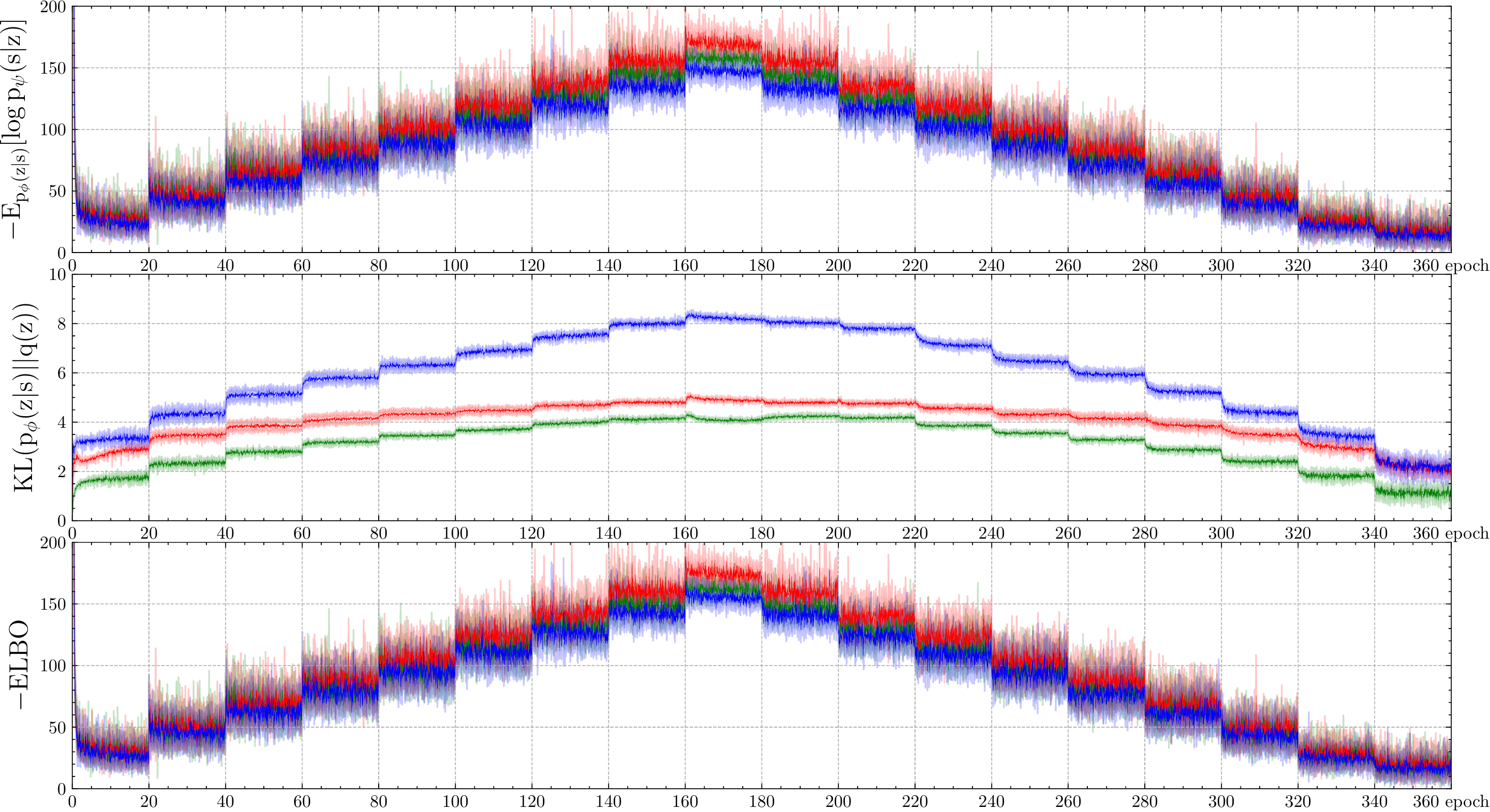}
    \caption{Experimental results for the MNIST dataset with changing diversity. 
    The plot shows values for negative $\beta$-ELBO,  $\mathrm{KL}[p_{\phi}(z|x)||q(z)]$, and $- \mathbb{E}_{p_{\phi}(z|x)} \log p_{\psi}(x|z)$ averaged across 6 different seeds. 
    Testing was conducted across three different combinations of $\beta$ and $z$ values: (\textcolor{forestgreen}{$\beta$=5,$z$=4}), (\textcolor{blue}{$\beta$=1,$z$=4}), (\textcolor{red}{$\beta$=1,$z$=2}).
    Note that when the diversity of the training samples does not change, the KL-divergence value converges to an equilibrium. 
    If the diversity of the samples increases, $\mathrm{KL}[p_{\phi}(z|x)||q(z)]$ also increases but  $\mathbb{E}_{p_{\phi}(z|x)} \log p_{\psi}(x|z)$ decreases and vice-versa. The $\{0, ..., 9\}$ digit set possesses the most diverse pixel value  configurations, as such it has the maximal negative ELBO value. 
    }
  \label{fig:explain_kl_change}
\end{figure}

In this experiment, we wish to show how the diversity of a dataset has a effect in the value of the negative ELBO. We begin by using two visual datasets: MNIST \cite{deng2012mnist} and Fashion-MNIST \cite{xiao2017fashion}. 
To this end, we control and iteratively increase the pixel value diversity of the dataset by constraining the digits presented to the system in a given number of epochs. 
Specifically, we begin by only allowing the set of digits $\{0, 1\}$, and then subsequently integrating one additional digit to the dataset every 20 epochs for training as follows: $\{0, 1, 2\}$, ..., $\{0, 1, ..., 9\}$.
Then, we remove digits in the following order: $\{0, ..., 8\}, ..., \{0, 1\}$. 
In doing so, the diversity of pixel values increases with new inclusions of digits and decreases with their removal.
Additionally, in order to control for dataset size effects, we make all the datasets to have the same number of images by adding blank ones in datasets with missing classes.
All those images will be fed into a VAE trained to maximize the ELBO in an unsupervised manner.
The results for the ELBO test phase are visualized in Fig. \ref{fig:explain_kl_change}. The plot shows the relationship between the negative $\mathrm{ELBO}$ and the diversity of the samples with different hyperparameters $\beta$ and $z$. Note how a change in the the value of negative ELBO is accompanied by a change in the degree of a sample's diversity thus corroborating our proof in Sec. \ref{sec:autotune_scaffolding_hyperparameters}.

\begin{figure*}[]
  \centering
  \begin{subfigure}{0.325\linewidth}
    \includegraphics[width=\linewidth]{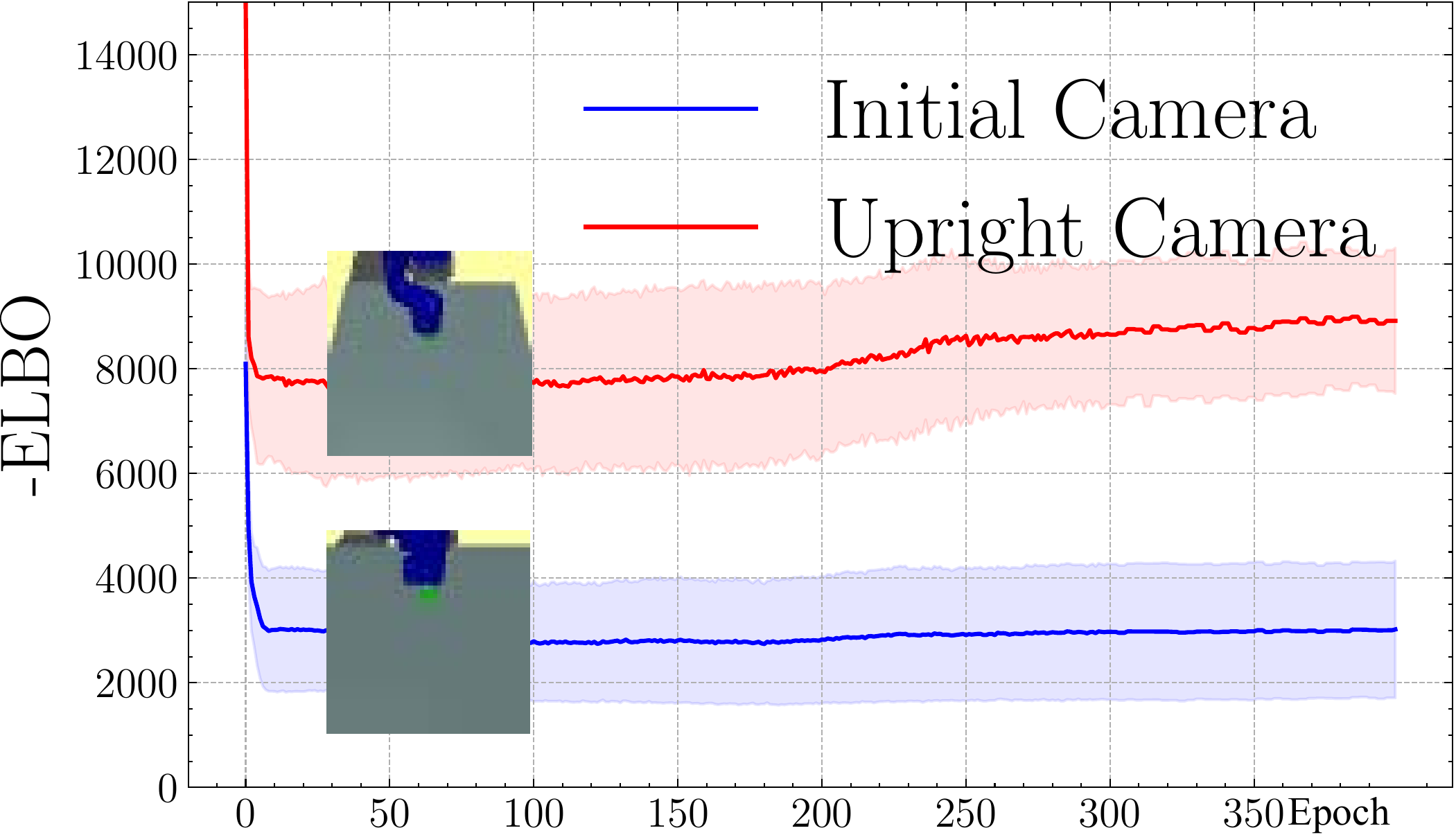}
    \caption{Sim:Angles}
  \end{subfigure}
  \begin{subfigure}{0.325\linewidth}
    \includegraphics[width=\linewidth]{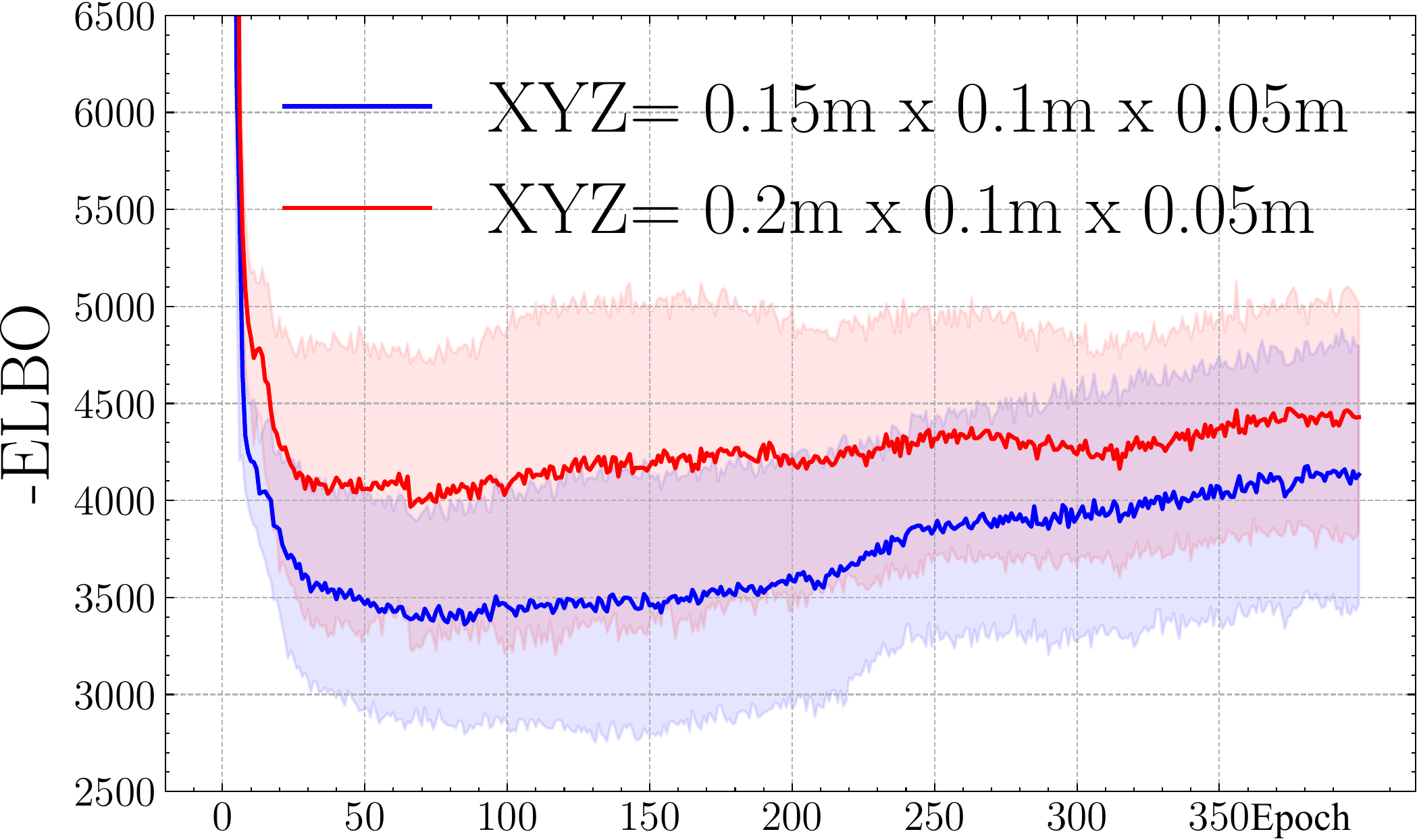}
    \caption{Sim:Workspace}
  \end{subfigure}
  \begin{subfigure}{0.325\linewidth}
    \includegraphics[width=\linewidth]{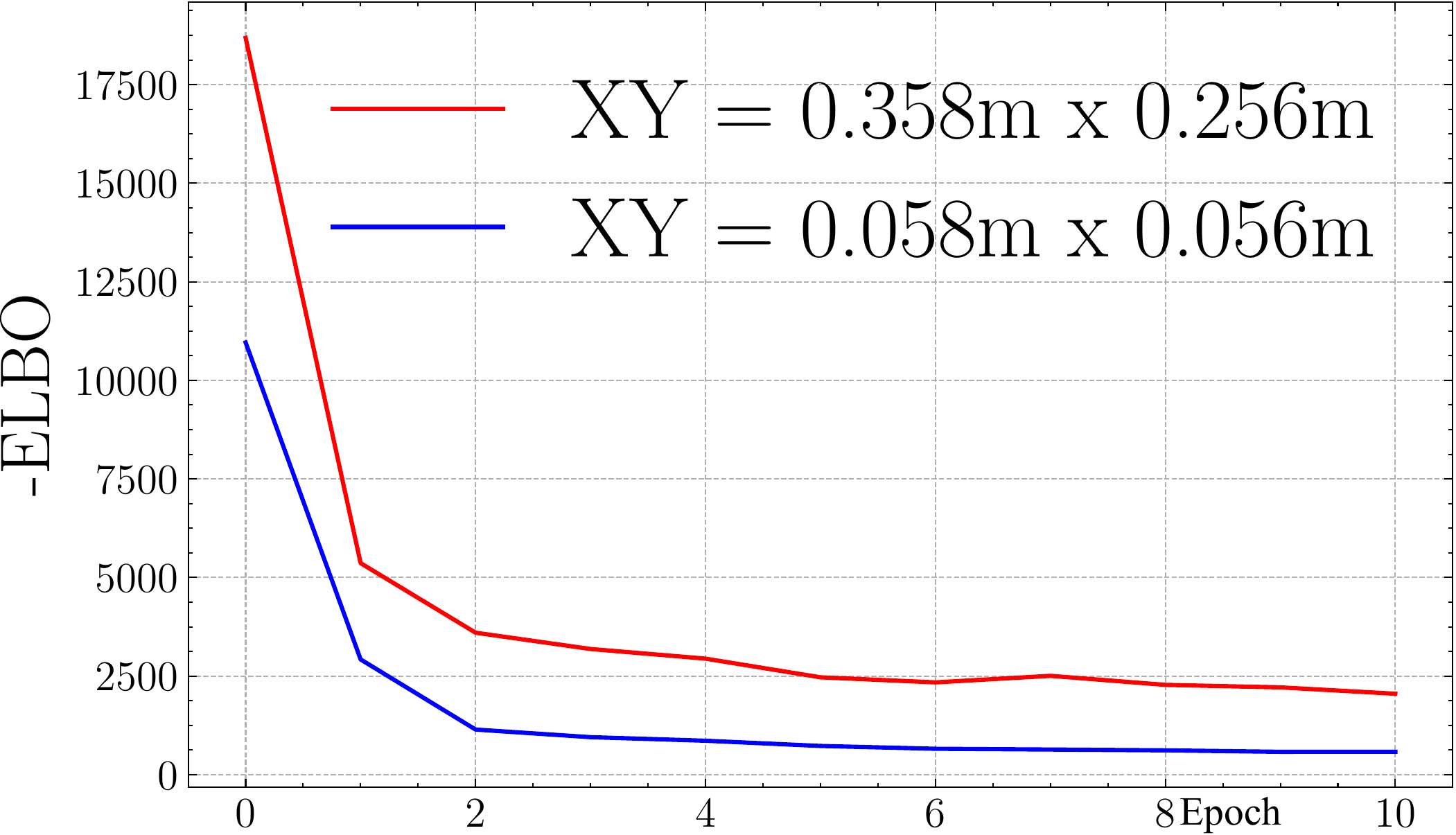}
    \caption{Real:Workspace}
  \end{subfigure}
    \caption{We show that the value of the negative $\beta-\mathrm{ELBO}$ in the visual robotic reach task in simulation (Sim) or real world (Real). In (a), negative $\beta-\mathrm{ELBO}$ increases if the camera captures more features of the robot arm. In (b) and (c), negative $\beta-\mathrm{ELBO}$ increases when the workspace is enlarged.
    }
  \label{fig:explain_elbo_on_RIG}
\end{figure*}

Secondly, we demonstrate this relationship in visual RL environments. 
To create different diversity of observations in the same environment, we setup the environment with different camera view angles and different workspaces of the robot arm's end-point.
In Fig. \ref{fig:explain_elbo_on_RIG}, when a larger workspace of robot arm and the upright camera view angle (can capture more movable features of the arm) both experience a larger value of negative $\mathrm{ELBO}$.

\subsection{Auto-tuning in Static Environment}
\label{exp:2_ablative}
In this section, we will compare the efficiency of our auto-tuning method with hyperparameter optimization.
We use Optuna \cite{akiba2019optuna} to optimize the three hyperparameters $\mathbf{N_e}, \mathbf{N_b}$ and $ \mathbf{N_\theta}$ in RIG and the single $\xi$ hyperparameter in our Auto-tuning with RIG.

\subsubsection{Experimental setup}
RIG is trained with Soft Actor-Critic \cite{pong2019skew} on \textit{Visual Push}, \textit{Visual Multi-Object Push}, \textit{No Wall 2D Navigation} and \textit{Multi-Wall 2D Navigation} environments (Fig. \ref{fig:screenshot_and_obs}).
During training, the agent's goal is to learn numerous skills from the environment.
During policy evaluation, the aim is to achieve a user-specified goal given an initial state. The user-specified goal $g$ can, for instance, be an image of the robot arm positioned some distance away from the puck.
For the robot experiments, a robotic arm is controlled using only RGB image observations. It does not have access to any ground truth reward as was done in the policy learning process of RIG (see Sec. \ref{sec:background} and Algo. \ref{alg:algorithm}).
Note that the VAE model is pre-trained with randomly collected samples in the environment before conducting reinforcement learning. The VAE model is subsequently fine-tuned during RL epochs.

If the transitions from a new exploration step along with the old transitions exceeded the replay buffer size, the excess trajectories will be placed as the earliest transitions in the replay buffer. If the replay buffer size decreases (with our method), the number of trajectories will be cut instead of the transitions. 

For Optuna, the given objective is to minimize the image distance during the evaluation phase with four workers and 40 trials.
The considered search space is the following: $\mathbf{N_e}, \mathbf{N_b} \in \{100, ..., 6000\}^2, \mathbf{N_{b}} \in \{3000, ..., 300000\}$ and $\xi \in [0.1 ; 2]$.

To highlight the necessity of selecting good values for those hyperparameters, we also compared with three baselines where some hyperparameter values are suboptimal:
\begin{itemize}
    \item Limited exploration steps with $\mathbf{N_e}= 100$, $\mathbf{N_b}= 300000$ and $\mathbf{N_{\theta}}= 6000$,
    \item Limited replay buffer size with $\mathbf{N_e}= 6000$, $\mathbf{N_b}= 3000$, $\mathbf{N_{\theta}}= 6000$,
    \item Limited gradient updates $\mathbf{N_e}= 6000$, $\mathbf{N_b}= 300000$, $\mathbf{N_{\theta}}= 100$.
\end{itemize}

\subsubsection{Results}

Among the search space, Optuna selected the following hyperparameters:
\begin{itemize}
    \item $\xi=1.142, \mathbf{N_e}=2852, \mathbf{N_b}=154375, \mathbf{N_\theta}=2799$ in \textit{Visual Push} environment,
    \item $\xi=1.218, \mathbf{N_e}=5645, \mathbf{N_b}=170937, \mathbf{N_\theta}=5978$ in \textit{Visual Multi-Object Push} environment,
    \item $\xi=0.608, \mathbf{N_e}=3432, \mathbf{N_b}=182087, \mathbf{N_\theta}=2943$ in \textit{No Wall 2D Navigation} environment,
    \item $\xi=0.763, \mathbf{N_e}=5770, \mathbf{N_b}=279702, \mathbf{N_\theta}=5700$ in \textit{Multi-Wall 2D Navigation} environment.
\end{itemize}
With those hyperparameters, we then average over ten runs per method and we report the mean $\pm$ standard deviation curves across metrics.
The performance metric is the image distance between the observation image and user-specified image in the evaluation phase. Instead of using task-relevant metric like puck distance in \textit{Visual Push} to do the evaluations, using image distance as metric to compare all the features between current image and user-specified image about both the puck's position and the hand's position is more comprehensive.

\begin{figure*}[]
  \centering
    \includegraphics[width=\linewidth]{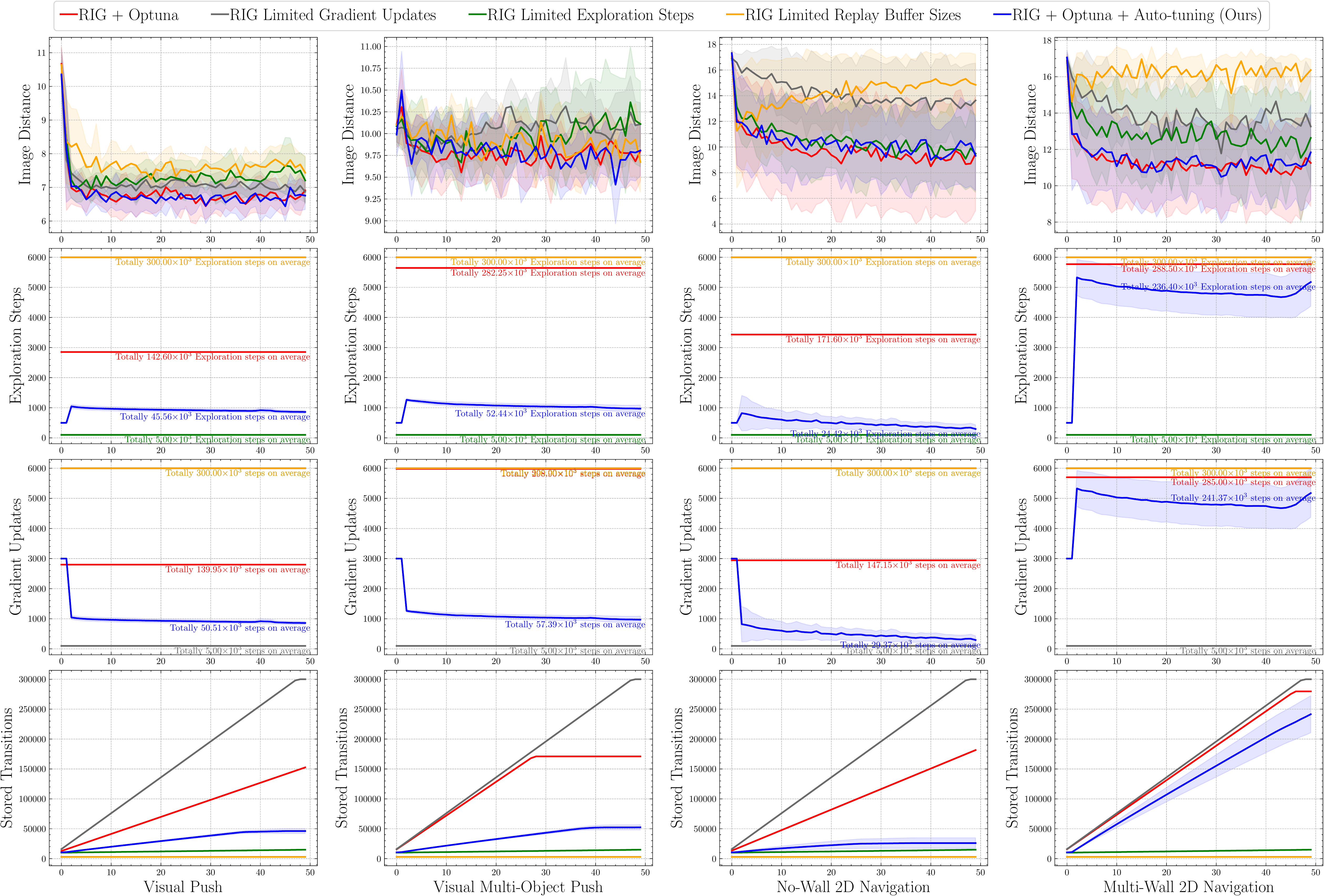}
    \\[0.5em]
    \caption{Comparison of hyperparameter optimisation (Optuna) with/without Auto-tuning and three suboptimal baselines.
    }
    \label{fig:ablation}
\end{figure*}

From Fig. \ref{fig:ablation}, we can see that the performances of the solutions found by Optuna with or without auto tuning are equivalent.
We also see that selecting suboptimal values for $\mathbf{N_{e,b,\theta}}$ in the three baselines may have a large negative impact on the performance.
Interestingly, our auto-tuning method always selects less gradient updates, exploration steps and stored transitions than the Optuna counterpart.
Our method is also capable of changing the values of the three underlying hyperparameters online, whereas it would require a much larger search space for Optuna to do the same.

\subsubsection{Interpretation}

As our auto-tuning method only have to tune one hyperparameter instead of three, we are naturally more sample efficient during the hyperparameter search.
As the final performance with or without auto-tuning is similar, it shows that our method does not impede the selection of the three underlying hyperparameter.
On the contrary, our method is able to find more interesting final solutions from a multi-objective point-of-view (performance and computational costs): it uses less gradient updates, exploration steps, and stored transitions for a similar performance.

In conclusion, in comparison with a direct hyperoptimization of $\mathbf{N_{e,b,\theta}}$, we need less interaction with the environment and we also improve the computational cost of the solution found.

%
%

\subsection{Coverage Performance of Auto-tuning}\label{subsec:coverage}
\begin{figure*}[]
  \centering
  \begin{subfigure}{0.49\linewidth}
    \centering
    \includegraphics[width=.96\linewidth]{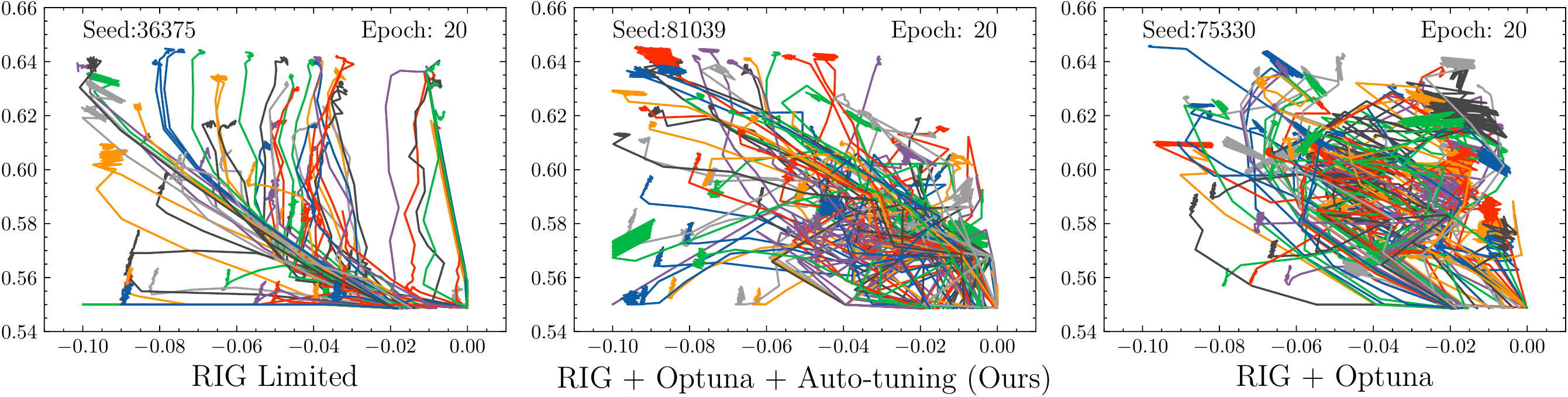}
    \caption{Trajectories from the evaluation phase at epoch 20}
  \end{subfigure}
  \begin{subfigure}{0.49\linewidth}
    \centering
    \includegraphics[width=.96\linewidth]{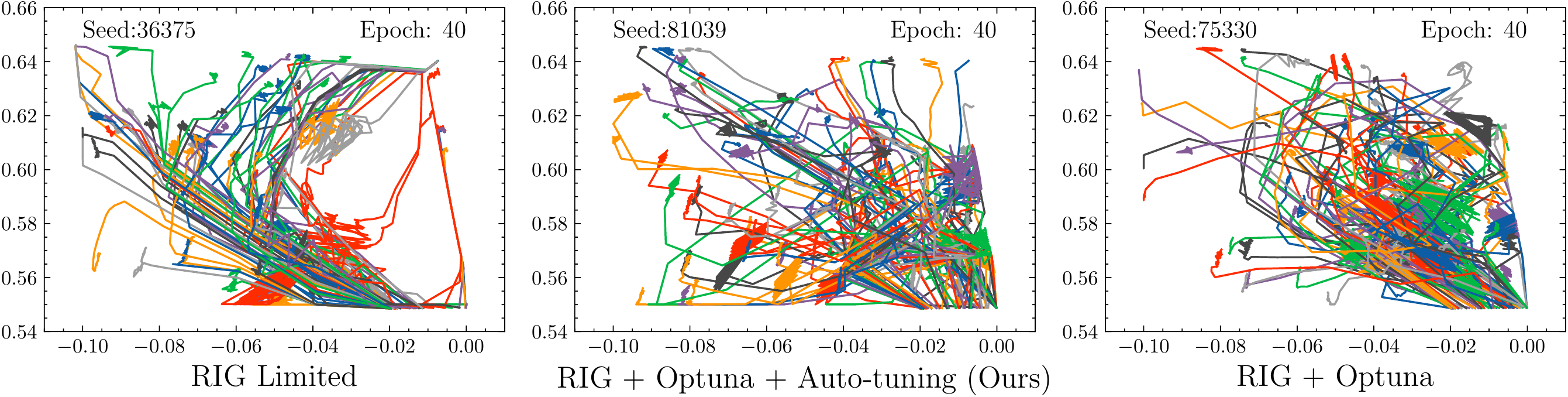}
    \caption{Trajectories from the evaluation phase at epoch 40}
  \end{subfigure}
    \caption{
    Comparing the coverage performance between RIG with Auto-tuning, RIG with optimal $\mathbf{N_{e,b,\theta}}$, and RIG with limited $\mathbf{N_{e,b,\theta}}$. Curves with different colours indicate different trajectories in the policy evaluation phase of the current epoch.}
  \label{fig:coverage_traj}
\end{figure*}

During the evaluation phase in RIG, the goals are a set of randomly collected images from the environment. 
Optimal values of $\mathbf{N_{e,b,\theta}}$ result in optimal result. 
Using optimal models to reach all the goals in the environment also implies that the agent achieves maximal coverage of the space.
The optimal result should finish all the possible goals in the environment, in other words, should achieve maximal coverage of the environment space.
To specifically visualize and compare the performances with different values $\mathbf{N_{e,b,\theta}}$.
We measure the coverage performance by plotting the end-effector trajectories during the policy evaluation phase. The environment we used to verified is \textit{Visual Push}, which need to learn how to push a puck and reach the desired goal of the end-effector.
In this experiment, we use the same hyperparameters values as mentioned in Sec. \ref{exp:2_ablative}. The space that the agent needs to cover is $xyz = 10 \times 10 \times 50~ cm^3$. The trajectories are shown in Fig. \ref{fig:coverage_traj} with $xy$ plane.
As for auto-tuning, either in epoch 20 or in epoch 40, a scaling parameter $\xi$=1.142 has qualitatively similar coverage of the environment as fixing the values $\mathbf{N_{e,b,\theta}}$ optimally.
Comparisons are also conducted under the limit hyperparameter value setup. In this case, it is difficult to cover the entire environment, resulting in less image distance as shown in Fig. \ref{fig:ablation}.

\subsection{Auto-tuning in Curriculum Setups}\label{subsec:single_domain_dynamic_change}
In this experiment, we further study auto-tuning in a curriculum learning environment. 
Under the curriculum setup, the agent does not know when the environment may change. Therefore, it is crucial to have an algorithm that can adapts its hyperparameters automatically after an environment change.
The curriculum learning environment is changed programmatically in simulation and manually in real-robot experiments.

We test the curriculum setup in two simulated tasks (2D navigation and robot manipulation) and one real robot manipulation task. Each change in the curriculum task essential introduces more visual diversity to the task.
First, for the simulated navigation task, we first use
a \textit{No-Wall 2D Navigation} environment, then a \textit{Multi-Wall 2D Navigation} environment is introduced at epoch 50, and finally a \textit{Multi-Color 2D Navigation} environment is inserted at epoch 100.
Second, for the simulated robot curriculum task, we begin with an object-less \textit{Visual Reach} environment, followed by the introduction of an object in the \textit{Visual Push} environment at epoch 100, followed by a \textit{Visual Multi-Object Push} environment at epoch 200. 
Finally, for the real-robot curriculum task, we begin with a \textit{Real Reach} environment, followed by a manual change of the camera angle view at epoch 20, followed by the manual addition of objects at epoch 30.

\begin{figure*}[]
  \centering
  \begin{subfigure}{.325\linewidth}
    \includegraphics[width=\linewidth]{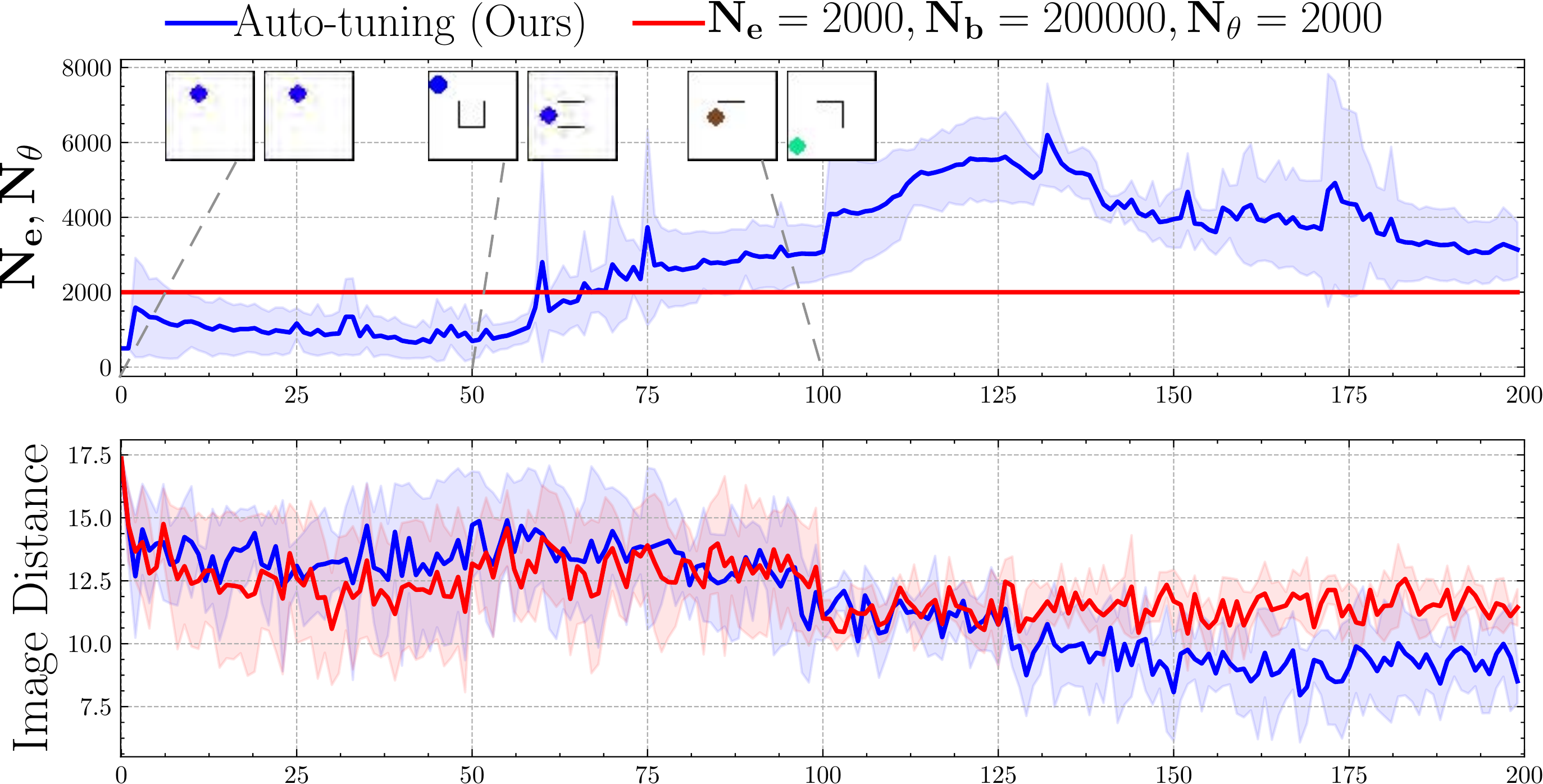}
    \caption{Navigation environment, $\xi = 0.608$}
  \end{subfigure}
  \begin{subfigure}{.325\linewidth}
    \includegraphics[width=\linewidth]{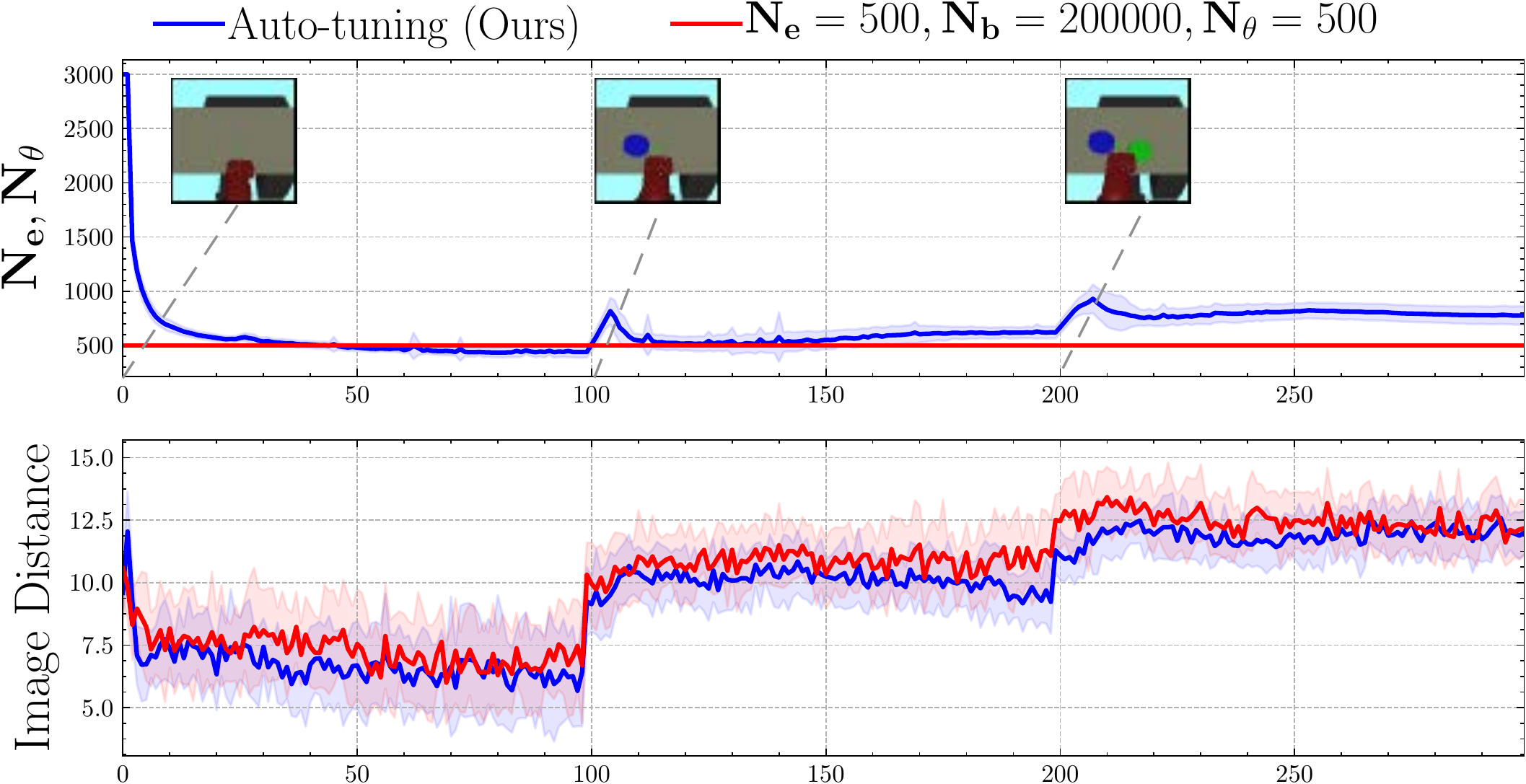}
    \caption{Robot environment, $\xi = 1$}
  \end{subfigure}
  \begin{subfigure}{.325\linewidth}
    \includegraphics[width=\linewidth]{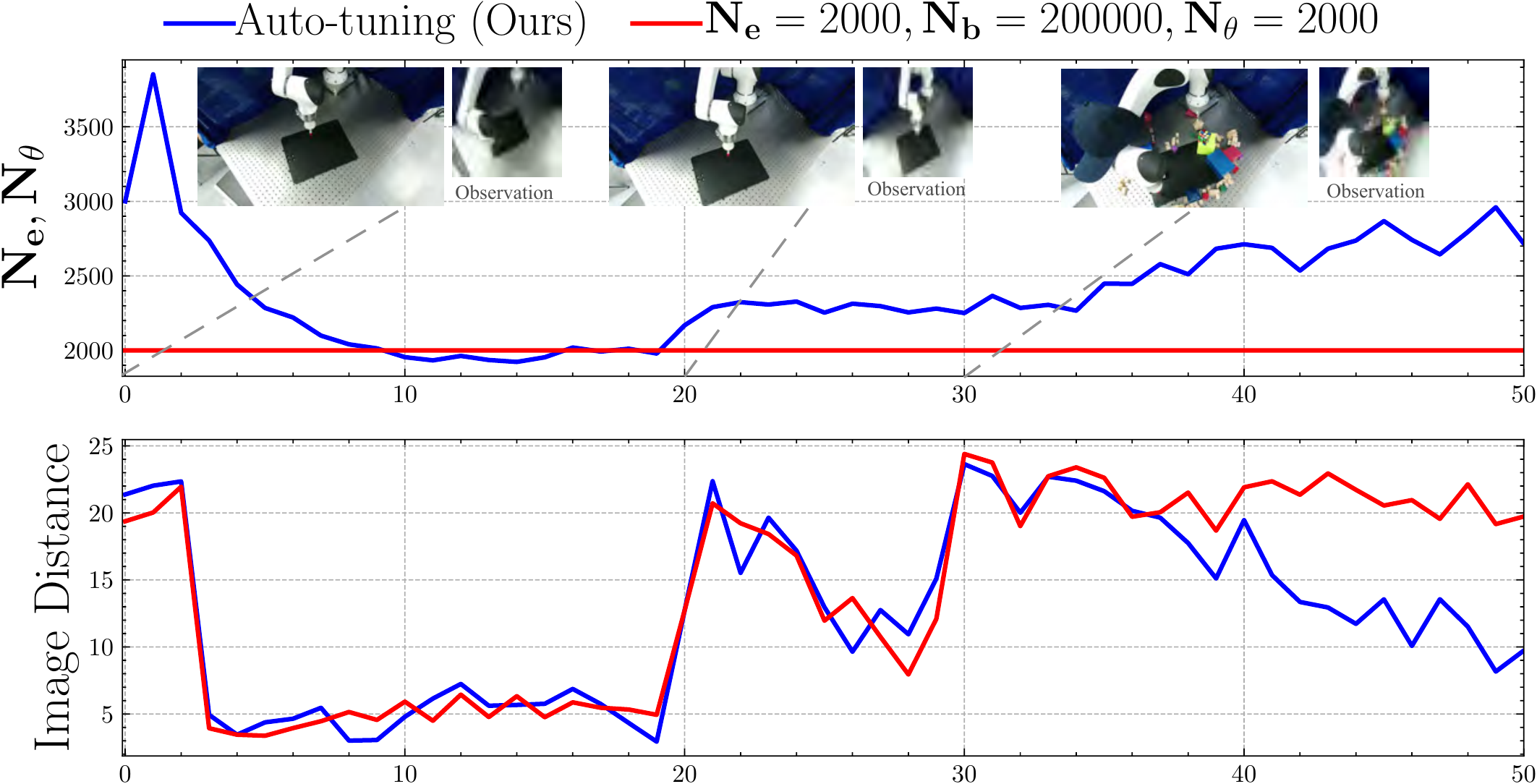}
    \caption{Real-world environment, $\xi = 1$}
  \end{subfigure}
    \caption{
    Performance comparison between RIG-SAC and Auto-tuning in different curriculum environments. We just plot the values of $\mathbf{N}_{\mathbf{e,\theta}}$. The values of $\mathbf{N}_{\mathbf{b}}$ is the product of the values of $\mathbf{N}_{\mathbf{e}}$ and the max path length $l = 50$.
    }
  \label{fig:curriculum}
\end{figure*}

As with previous experiments, the negative $\mathrm{ELBO}$ automatically changes in response to modifications in the environment prompted by the curriculum as shown in Fig. \ref{fig:curriculum}. 
Hence, with Auto-tuning, an agent can adapt its values of $\mathbf{N}_{\mathbf{e,b,\theta}}$ to optimize learning.
Note that in RIG, with fixed hyperparameter values $\mathbf{N}_{\mathbf{e,b,\theta}}$, the agent is still able to learn optimally in the first simulated robot curriculum setup; however, in the other two curriculum examples presented, the learning becomes suboptimal. 

In conclusion, the values $\mathbf{N}_{\mathbf{e,b,\theta}}$ are calculated directly with the negative $\mathrm{ELBO}$ each epoch. 
Auto-tuning increases the values of $\mathbf{N}_{\mathbf{e,b,\theta}}$ after the environment becomes more diverse, and learning is faster than RIG with fixed values.

\section{Related Work}\label{sec:related work}

Self-supervised RL learns skills autonomously and incrementally \cite{baranes2013active, pathak2019self, WardeFarley2019UnsupervisedCT}.
In self-supervised RL, the goal generator can be trained online  \cite{florensa2018automatic,nair2018visual,pong2019skew} or offline \cite{nair19ccrig}. Our work studies online methods as offline methods are unable to recognize dynamic environment changes in future episodes.
Also, self-generated goal agents can be easily combined with modern RL algorithms like PPO \cite{florensa2018automatic}, TD3 \cite{nair2018visual}, and SAC \cite{pong2019skew}.
Nevertheless, previous works have fixed the values of $\mathbf{N}_{\mathbf{e,b,\theta}}$ in off-policy learning regardless of how diverse the environments.

Goal generators like VAEs are trained in an unsupervised manner \cite{diederik2014auto,higgins2017beta}, which allows to easily apply them in RL. Moreover, VAEs can be used for exploration \cite{klissarovvariational_workshop,kim2019curiosity}. Rarely is the ELBO used in VAEs to auto-tune hyperparameters.
VAEs can use information theory for estimation and optimization \cite{chen2016infogan,poole2019variational,zhao2019infovae}.
However, previous works only analysed datasets with fixed diversity whose sample entropy did not change over time \cite{diederik2014auto,nair2018visual}.
Instead, we estimate how a dataset's changing diversity affects the ELBO of the VAEs.

The effectiveness of hyperparameter optimization depends on parameter-sampling algorithms like TPE \cite{bergstra2011algorithms}, GP-BO \cite{shahriari2015taking}, and Autotune \cite{koch2018autotune}. However, automatic online tuning of hyperparameters is studied less.
Rather than tuning hyperparameters online, the values of hyperparameters are typically fixed since the diversity of the samples does not change \cite{nair2018visual}, and few studies have learned to optimize these valus online. 
However, such fixed parameterization is ineffective in learning tasks where the diversity of the environment changes online. 
There has been much work in deterministic scheduling: some decay the learning rate of gradient descent at every epoch \cite{Goodfellow2015DeepL}, 
Pong et al. \cite{pong2019skew} decrease the number of VAE iterations according to a schedule. 
These methods empirically derive the schedule rather than learning it according to the varying diversity of an environment.
Recently, to tune hyperparameters automatically, Fakoor et al. \cite{fakoor2020p3o} used a normalized effective sample size to tune the IS clipping threshold and the KL regularization coefficient.
Zahavy et al. \cite{zahavy2020self} made progress on self-tuning all the differentiable hyperparameters of an actor-critic loss function by using meta-gradients. 

\section{Conclusion}\label{sec:discussion}
Hyperparameter auto-tuning in self-supervised robot learning makes RIG easier to apply in practice to different domains with the auto-tuning of three hyperparameters $\mathbf{N_{e,b,\theta}}$ instead of relying on costly hyperparameter optimization.
Hyperparameters $\mathbf{N_{e,b,\theta}}$ are common in various RL framework, so our proposition can be extended to various VAE-based RL algorithms.
To choose adequate hyperparameter values $\mathbf{N}_{\mathbf{e,b,\theta}}$, we estimated and verified that the negative $\mathrm{ELBO}$ (loss function of VAEs) has a lower bound related to the diversity of the training dataset. 
Moreover, the ELBO varies according to the change of diversity, which is so far un-studied.
A nice property of our method is that it does not require much additional computational cost, since it directly uses the metric already computed during the VAE evaluation phase in RIG.
Experiments demonstrated that auto-tuning can save time and resources by adaptively selecting $\mathbf{N}_{\mathbf{e,b,\theta}}$ according to the diversity of goals of a given task.
Designing deep RL algorithms with hyperparameters that do not need to be hand-tuned is an important step to make RL more practical.
Future investigation to auto-tuning other hyperparameters would be worthwhile. 



\bibliographystyle{IEEEtran}
\bibliography{IEEEabrv,references}
\end{document}